\documentclass[journal,letterpaper]{IEEEtran}
\IEEEoverridecommandlockouts
\usepackage[table]{xcolor}
\usepackage{amsmath,amssymb,amsfonts}
\newtheorem{defi}{Definition}
\usepackage{algorithm}
\usepackage{algorithmic}
\usepackage{graphicx}
\usepackage[caption=false,font=footnotesize]{subfig}  
\usepackage{textcomp}
\usepackage{xcolor}
\usepackage{cite} 
\usepackage{multirow}
\usepackage{float} 
\usepackage{url}

\def\BibTeX{{\rm B\kern-.05em{\sc i\kern-.025em b}\kern-.08em
    T\kern-.1667em\lower.7ex\hbox{E}\kern-.125emX}}

\usepackage{booktabs}
\usepackage[table]{xcolor}
\usepackage{hyperref}
\hypersetup{
    colorlinks=true,
    linkcolor=blue,
    filecolor=blue,
    urlcolor=blue,
    citecolor=blue
}

\begin{document}
\title{PSSI-MaxST: An Efficient Pixel-Segment Similarity Index Using Intensity and Smoothness Features for Maximum Spanning Tree Based Segmentation\\

\thanks{\textit{*The authors are with Department of Computer Science and Engineering, Visvesvaraya National Institute of Technology Nagpur 440010, India.
Email: sshejole132@gmail.com, gauravmishra@cse.vnit.ac.in}
}
}

\author{Kaustubh~Shivshankar~Shejole
        and Gaurav~Mishra%
}

\maketitle

\begin{abstract}

Interactive graph-based segmentation methods partition an image into foreground and background regions with the aid of user inputs. However, existing approaches often suffer from high computational costs, sensitivity to user interactions, and degraded performance when the foreground and background share similar color distributions. A key factor influencing segmentation performance is the similarity measure used for assigning edge weights in the graph.
To address these challenges, we propose a novel Pixel Segment Similarity Index (PSSI), which leverages the harmonic mean of inter-channel similarities by incorporating both pixel intensity and spatial smoothness features. The harmonic mean effectively penalizes dissimilarities in any individual channel, enhancing robustness. The computational complexity of PSSI is $\mathcal{O}(B)$, where $B$ denotes the number of histogram bins.
Our segmentation framework begins with low-level segmentation using MeanShift, which effectively captures color, texture, and segment shape. Based on the resulting pixel segments, we construct a pixel-segment graph with edge weights determined by PSSI. For partitioning, we employ the Maximum Spanning Tree (MaxST), which captures strongly connected local neighborhoods beneficial for precise segmentation.
The integration of the proposed PSSI, MeanShift, and MaxST allows our method to jointly capture color similarity, smoothness, texture, shape, and strong local connectivity. Experimental evaluations on the GrabCut and Images250 datasets demonstrate that our method consistently outperforms current graph-based interactive segmentation methods such as AMOE, OneCut, and SSNCut in terms of segmentation quality, as measured by Jaccard Index (IoU), $F_1$ score, execution time and Mean Error (ME).
Code is publicly available at:
\url{https://github.com/KaustubhShejole/PSSI-MaxST}

\end{abstract}


\section{Introduction}


Image segmentation partitions an image into distinct regions. It serves as a crucial preprocessing step for tasks such as object tracking, recognition, and various computer‑vision applications. A variety of techniques are available, including thresholding, edge detection, region‑based methods, watershed algorithms, and energy‑based approaches such as GraphCut \cite{6223428,Blake1987VisualR}. Graph‑based segmentation methods covering graph‑cut, interactive, minimum spanning tree (MST)‑based, and pyramid‑based approaches are surveyed in \cite{camilus2012review,peng2013survey}.

GrabCut \cite{rother2004grabcut} iteratively builds an appearance model using the Gaussian Mixture Model (GMM) inside a user‑provided bounding box. This iterative refinement makes it relatively slow. To address this inefficiency, Tang et al. \cite{tang2013grabcut} proposed OneCut, which replaces the log‑likelihood term with an explicit $L_{1}$ distance between the object and background appearance models. OneCut then performs a single global optimization via graph cuts (GC), reducing the computational complexity while maintaining good segmentation results.

However, for some images, this method was found to be prone to producing isolated regions \cite{qu2022interactive}. Because its energy term grows with the number of boundary edges and relies solely on an $L_{1}$ appearance distance, OneCut ignores spatial coherence. Moreover, OneCut is highly sensitive to initial seed placement, and our experiments show that this seed‑sensitivity often leads to sub‑optimal segmentations.


\cite{qu2022interactive} proposed Image Segmentation based on Appearance Model and Orientation Energy (AMOE), which extends OneCut by incorporating both pixel‑to‑pixel distance and brightness cues derived from intervening contours. AMOE achieves higher segmentation accuracy by optimizing this combined energy.

However, in images where foreground and background exhibit similar appearance, AMOE can erroneously label foreground as background (and vice versa), producing fragmented or incomplete regions. Furthermore, because AMOE still models every pixel as a graph node, it inherits OneCut’s high computational complexity and sensitivity to initial seeds. 
Experiments confirm that when user‑provided seeds are sparse, both AMOE and OneCut often yield sub‑optimal segmentations, limiting their practical usability.


Semi‑Supervised Normalized Cuts (SSNCut) \cite{chew2015semi} first oversegments the image using Simple Linear Iterative Clustering (SLIC). It then constructs a graph in which each superpixel (i.e., pixel-segment) is represented as a vertex. SSNCut assigns edge weights using the globalized probability of boundary (gPb) computed at each pixel \cite{arbelaez2010contour}. The runtime of SSNCut increases due to higher complexity of calculating gPb. 
Experiments demonstrate that SSNCut can lead to poor segmentation results under weak boundary cues, and it remains sensitive to initial seed placement \cite{qu2022interactive}.

With the motivation of addressing the limitations of the above methods, we propose a novel and efficient Pixel Segment Similarity Index (PSSI) for assigning edge-weights in pixel-segment graph in our improved interactive and iterative graph-based image segmentation method. The proposed PSSI metric computes the harmonic mean of inter-channel similarities by incorporating both pixel intensity and spatial smoothness features. The harmonic mean formulation penalizes dissimilarity in any individual channel, thereby promoting consistency across all channels.The computational complexity of PSSI is $\mathcal{O(B)}$, improving upon the $\mathcal{O}(\mathcal{B}^3)$ cost of Bhattacharyya-style measures \cite{1573105974836363008} used previously for getting similarity between pixel-segments \cite{ning2010interactive}, where $\mathcal{B}$ refers to the number of bins. 

Our segmentation framework begins with low-level segmentation using MeanShift, which effectively captures color, texture, and segment shape. Based on the resulting pixel segments, we construct a pixel-segment graph with edge weights determined by PSSI. 
For partitioning, we employ the Maximum Spanning Tree (MaxST), which captures strongly connected local neighborhoods beneficial for precise segmentation.
The integration of the proposed PSSI, MeanShift, and MaxST allows our method to jointly capture color similarity, smoothness, texture, shape, and strong local connectivity.


The proposed framework (PSSI–MaxST) when evaluated on the GrabCut \cite{rother2004grabcut} and Images250 \cite{qu2022interactive} benchmark datasets, outperforms existing graph-based interactive image segmentation methods such as OneCut \cite{tang2013grabcut}, AMOE \cite{qu2022interactive}, and SSNCut \cite{chew2015semi} in terms of segmentation quality, as measured by Jaccard Index (IoU), $F_1$ score, execution time and Mean Error (ME). It results in better segmentation results while avoiding isolated regions and handling cases where foreground and background share similar appearance. Our method requires only a few seed points and gives reliable results under weak boundary cues, since PSSI does not depend on explicit boundary information.


The paper is organized as follows: Section \ref{related_work} describes the related work. Section \ref{proposed_method} discusses the proposed method. Section \ref{experimental_analysis} presents the experimental analysis of our work. In Section \ref{Conclusion}, we conclude our work with some future research directions.

\section{Related Work}\label{related_work}


Interactive image segmentation refers to foreground-background extracted guided by user inputs. Ramadan et al. (2020) \cite{ramadan2020survey} surveys recent interactive image segmentation methods. Graph-based methods, particularly graph cut and minimum spanning tree (MST) algorithms, have been widely applied to clustering \cite{xue2024comprehensive} and interactive image segmentation \cite{peng2013survey} Wu and Leahy first introduced pixel-graph cuts for segmentation in 1990 \cite{wu1990tissue}, and Shi and Malik later formalized normalized cuts (Ncut) as a global energy minimization framework \cite{shi2000normalized,6223428}. Various techniques for image segmentation are covered in \cite{mortensen1998interactive, falcao2000ultra, falcao1998user, braquelaire1998image, 937505, boykov2006graph}. 
Building on these foundations, later work focused on efficiency, richer energy terms, and graph-size reduction. GrabCut, OneCut, and AMOE perform binary partitioning via graph cuts on the full pixel graph \cite{rother2004grabcut,tang2013grabcut,qu2022interactive}, while SSNCut reduces graph size by clustering pixels into superpixel nodes and assigning edge weights using globalized probability of boundary (gPb) \cite{chew2015semi}.


Minimum spanning tree (MST)–based algorithms offer an alternative graph‑based approach to image segmentation, leveraging cluster theory to extract coherent regions. Zahn introduced an MST‑based clustering method in 1971, constructing an MST over the data points and removing inconsistent edges to form connected components that represent clusters \cite{zahn1971graph,camilus2012review}. Xu et al.\ \cite{xu1998segmentation,xu19972d} proposed a tree‑partitioning algorithm that splits the MST built from image pixels into sub‑trees, each containing a minimum number of vertices and exhibiting significantly different average gray levels from its neighbors. However, noise can distort the MST structure, causing objects to fragment across multiple sub‑trees and degrading segmentation quality \cite{camilus2012review}. MST‑based neighborhood graphs have since proven useful in various clustering and partitioning applications \cite{9044397}. Building on these clustering ideas, He et al. (2008)\ \cite{he2008new} later showed that image segmentation can be posed as an optimization problem of finding the maximum spanning tree over a pixel similarity graph, with pixels as vertices and pairwise similarities as edge weights.


To reduce computational complexity and improve structural coherence, many segmentation pipelines begin with low-level segmentation techniques. Low-level segmentation methods like SLIC \cite{item_2dd26d473d0043eb9e314610db94a26e} and Mean Shift \cite{zhou2011mean} are commonly used as pre-processing steps for interactive segmentation, significantly reducing graph size compared to pixel-level formulations \cite{chew2015semi,ning2010interactive}. Ning et al. \cite{ning2010interactive} construct a superpixel graph using Bhattacharyya similarity measure \cite{1573105974836363008} to compute edge weights between SLIC superpixels. In SSNCut \cite{chew2015semi}, the globalized probability of boundary (gPb) \cite{arbelaez2010contour} at each pixel is used, the maximum gPb across eight orientation angles is computed, and the weight between superpixels \(i\) and \(j\) is defined as:

\[
W_{i,j} =
\exp( \max\{ \text{gPb}(\boldsymbol{p}_{i,j})/ {\rho}) 
\] if $p_{i,j} \leq r$ and 
$W_{i,j} = 0$, otherwise, where \(||\boldsymbol{p}_{i,j}||\) is the line segment connecting the centroids of superpixels \(i\) and \(j\), and \(\rho\) is a constant. We set \(r\) to be 0.1 times the maximum of the image width and height, and \(\rho = 0.1\).




While prior methods such as SSNCut \cite{chew2015semi} try to reduce complexity through low-level segmentation and apply graph cuts over superpixels, they rely on boundary-based similarity cues that result in higher runtime and can lead to poor performance when boundaries are weak \cite{qu2022interactive}. Similarly, the Bhattacharyya similarity used in \cite{ning2010interactive} is computationally expensive, with complexity cubic in the number of histogram bins. For overcoming these limitations, we propose a novel similarity measure (PSSI) that is both agnostic to edge cues and is linear in the number of bins, enabling efficient and effective weight assignment in pixel-segment graphs facilitating effective segmentation via MaxST-based partitioning.

\section{Proposed Approach}
\label{proposed_method}



Graph-based segmentation techniques often suffer from high computational costs and limited segmentation accuracy, particularly when operating directly at the pixel level. To address this, low-level segmentation methods such as SLIC and Mean-Shift are commonly used to group pixels into pixel segments, thereby reducing computational complexity. However, these pixel-segments are typically over-segmented and lack semantic meaning, making it essential to define an effective similarity measure between them. We propose a novel and efficient similarity measure for computing edge weights in a pixel-segment graph, as detailed in Section \ref{proposed_similarity_measure}. Our approach begins with Mean-Shift segmentation to generate pixel-segments, which form the nodes of the graph. Users provide sparse scribbles marking foreground and background regions, with an optional input  of bounding box, which are used to label the corresponding segments. A maximum spanning tree is then constructed over this graph using the proposed similarity weights, and segmentation is obtained via iterative MaxST-based partitioning. This process enables efficient and accurate object segmentation by leveraging both user input and global image structure.

\begin{figure}[ht] 
    \centering 
    \includegraphics[width=0.4\textwidth] {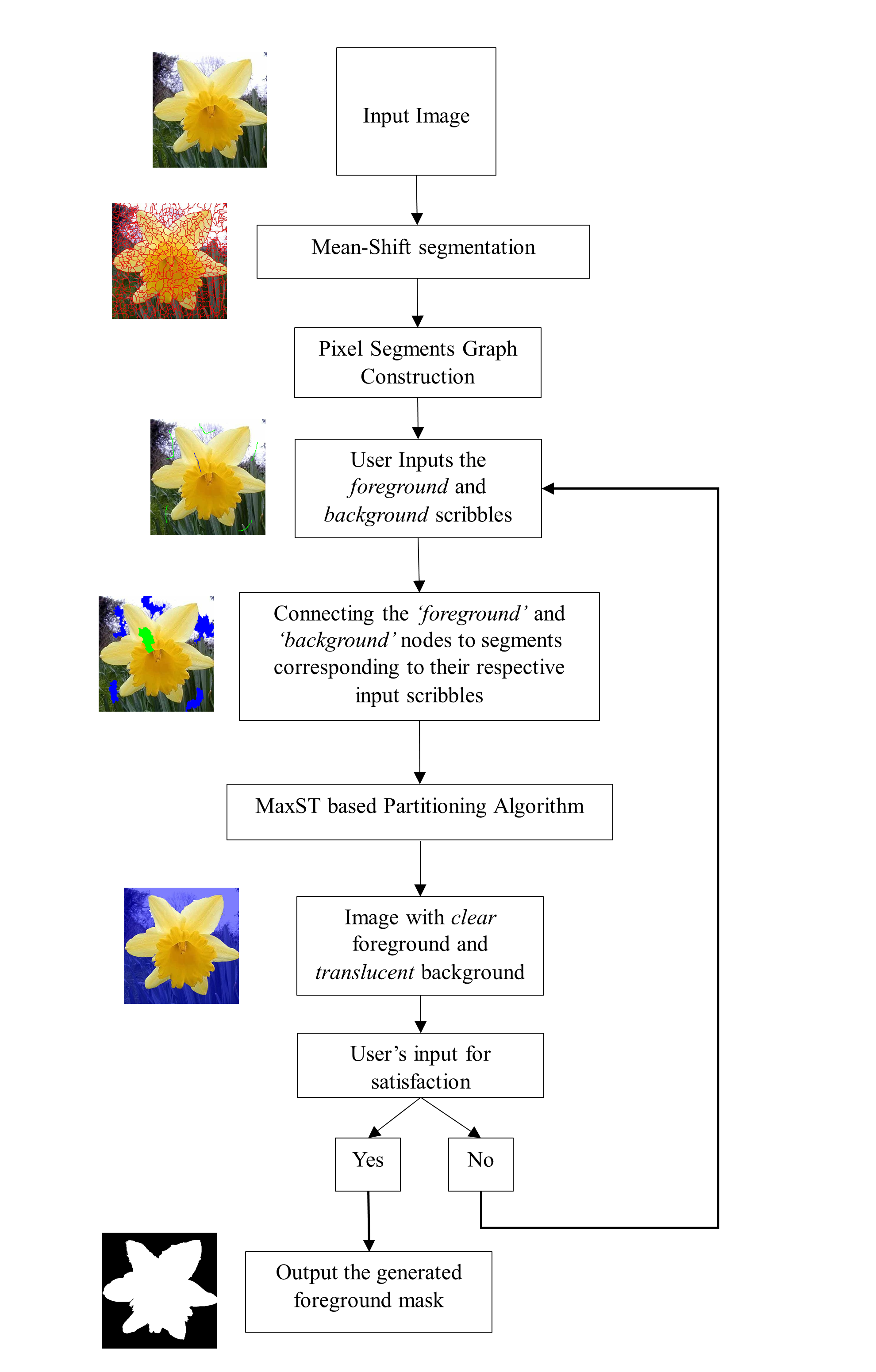} 
    \caption{Flow chart of the proposed algorithm.}
    \label{fig:flow chart} 
\end{figure}

Fig. \ref{fig:flow chart} represents  the flow chart of the proposed algorithm. 

The details of the proposed method are presented in the following subsections.

\subsection{Low-level segmentation}
\label{low_level_segmentation}
Low-level segmentation techniques serve as essential foundations for a variety of applications, particularly in the domain of image segmentation. These methods enable the effective partitioning of images into some regions which can facilitate further analysis and interpretation. There are several popular low-level segmentation methods such as SLIC \cite{item_2dd26d473d0043eb9e314610db94a26e}, Mean-Shift \cite{zhou2011mean}, etc.
SLIC creates the superpixels of about the same shape and dimensions whereas Mean-Shift creates pixel-segments that represent uniform color and an associated geometric shape. Thus, logically we can say that Mean-Shift is helpful in image segmentation as the segments are conforming to the geometric shape.

\subsection{Getting foreground and background pixel-segments}
\label{getting_user_validated_pixel_segments}
To effectively utilize the user's knowledge of foreground and background, we require the user to scribble foreground and background pixels. Using both foreground and background markers helps address weak boundaries by adding more seeds along the boundary \cite{937505, boykov2006graph}. After obtaining the marked pixels (from scribbles), we derive two corresponding lists for foreground and background pixel-segments. These lists serve as the initial seeds for the segmentation process.
In addition with these scribbles, bounding box can also be used to label all the pixels outside bounding box as strictly belonging to background.


\subsection{Graph Construction}
\label{graph-construction}

For constructing the graph, the $N$ pixel-segments generated from Mean-Shift segmentation  serve as the main nodes in the graph $G$. Additionally, we include two special nodes representing the `background' and `foreground'. Hence, we have $N+2$ nodes in a graph.
There are two main steps required for the construction of the graph. In the first step, we establish connections between the pixel-segments that the user has marked as foreground or background  which reflects the user's understanding of the image's content to terminal nodes.  After this, we get neighboring pixel-segments by taking spatial adjacency into account. The edges in the graph between pixel-segments are assigned the weights as given by the similarity measure (PSSI) detailed in the following section. Table \ref{tab:summary_edges_weights} summarizes the edge-weight assignments to pixel-segments.

\begin{table}[!t]
\caption{Calculation of edge weights}
\label{tab:summary_edges_weights}
\centering
\renewcommand{\arraystretch}{1.2}
\begin{tabular}{lll}
\toprule
\textbf{Edge} & \textbf{Weight} & \textbf{For} \\
\midrule
$\{p, q\}$ & $PSSI(p,q)$ & $p$ and $q$ are neighbors\\
$\{p, q\}$ & $0$ & $p$ and $q$ are not neighbors\\
$\{p, S\}$ & $\mathit{INFTY}$ & $p \in \mathit{Foreground \text{ }Scribbled}$ \\
$\{p, T\}$ & $\mathit{INFTY}$ & $p \in \mathit{Background \text{ }Scribbled}$ \\
\bottomrule
\end{tabular}
\end{table}

\subsection{PSSI: An Efficient Pixel Segment Similarity Index}
\label{proposed_similarity_measure}

To assign weights to edges between neighboring pixel-segments, one must choose an appropriate similarity measure. A popular choice in the literature is the Bhattacharyya coefficient, which treats two normalized histograms as vectors and computes the cosine of the angle between them \cite{1573105974836363008,derpanis2008bhattacharyya}:
\begin{equation}
  \rho(R,Q) \;=\; \sum_{u=1}^{B^3} \sqrt{\mathrm{Hist}_R^u \,\cdot\, \mathrm{Hist}_Q^u}\,,
\end{equation}
where $\mathrm{Hist}_R$ and $\mathrm{Hist}_Q$ are the normalized histograms of segments $R$ and $Q$, the superscript $u$ indexes the $u$‑th bin, and $B$ is the number of bins per channel. Its computational complexity is $\mathcal{O}(B^3)$, which can become expensive for large $B$.

To reduce the computational complexity of the existing similarity measure, we handled each color space separately, observing that the intra-segments usually have a gradual gradient or change in color and hence we must take neighborhood information into consideration while keeping in hand the cosine similarity between two histograms. We begin by defining per‑channel normalized histograms:

\begin{defi} \textbf{(Normalized Histogram).} The Normalized Histogram is defined as a histogram representing the relative frequency of each intensity level in a pixel-segment, normalized to sum to one. This can be mathematically expressed as:

\begin{equation}
    H_{\text{norm}}(i) = \frac{H(i)}{N}
\label{eq H-norm}
\end{equation}
where \(H(i)\) denotes the count of occurrences of intensity level \(i\) in the pixel-segment, and \(N\) being the total number of pixels in the pixel-segment.

\end{defi}
        

     \begin{defi}
     \label{ihsi_defi_label}
     \textbf{(Inter-Histogram Similarity Index)} We define the Inter-Histogram Similarity Index (IHSI), an index measuring the degree of similarity between two histograms, as follows:

        \begin{equation}
        \label{eq IHSI}
        \text{IHSI}(P, Q) = \sqrt{P^\top Q + \lambda \cdot P^\top A Q}
        \end{equation}
        where \(P, Q \in \mathbb{R}^B\) are normalized histograms, and \(A \in \mathbb{R}^{B \times B}\) is a a tridiagonal \emph{neighborhood adjacency matrix} encoding 1-bin adjacency defined as:
        \[
        A_{i,j} =
        \begin{cases}
        1, & \text{if } |i - j| = 1 \\
        0, & \text{otherwise}
        \end{cases}
        \]
        Here, \(B\) denotes the total number of bins, and the hyperparameter \(\lambda\) controls the importance of neighborhood similarity.
        We set the hyperparameter the neighborhood importance factor $\lambda$ to 0.2 for our experiments.  Although our method shows strong robustness to the value of \(\lambda\) (with stable performance in the range \([0.15, 0.25]\)), we retain it as a tunable parameter because it captures an important trade-off between appearance similarity and neighborhood consistency. This makes the formulation more interpretable and adaptable across datasets with varying texture or boundary characteristics. A higher \(\lambda\) gives greater importance to the neighborhood bins. We specifically apply the square root to each channel-wise similarity score because the square root function is concave over the interval $[0,1]$. This transformation increases low values more than high ones, effectively stretching the lower end of the similarity range. Such behavior is analogous to gamma correction with $\gamma = 0.5$, and it enhances perceptual sensitivity by amplifying subtle similarities while compressing larger ones. As a result, the similarity metric becomes more uniform across channels and better aligned with human visual perception, particularly in cases of color imbalance or low-contrast regions.
        \end{defi}



        We can see that in \eqref{eq IHSI} $P^\top Q$ considers the cosine similarity between two bins and $\lambda \cdot P^\top A Q$ term takes into consideration the relationship of smooth deviation with neighboring bins where $\lambda$ controls the weightage given to the consideration. Hence more similar the bin vectors of two colors, more value of the similarity index. 
        
        The similarity measure between two pixel-segments is computed by first extracting the segments \(S_P\) and \(S_Q\) from their labels \(L_P\) and \(L_Q\), respectively.  
        Next, we form normalized histograms \(P\) and \(Q\) for each of the RGB channels of \(S_P\) and \(S_Q\) using \eqref{eq H-norm}.  We then compute per‐channel similarity indices \(IHSI_R\), \(IHSI_G\), and \(IHSI_B\) via \eqref{eq IHSI} applied to \((P_R,Q_R)\), \((P_G,Q_G)\), and \((P_B,Q_B)\).  Finally, the overall similarity index \(PSSI\) is obtained as the harmonic mean of square roots of these three channel‐wise similarity indices i.e.,
\begin{equation}
    PSSI(P, Q) \;=\; HM\bigl(IHSI_R,\;IHSI_G,\;IHSI_B\bigr)
    \label{eq:PSSI}
\end{equation}
  
Its computational complexity is $\mathcal{O}(B)$, where $B$ refers to number of bins per channel, making it computationally efficient.

\subsection{ Maximum spanning tree based graph partitioning} \label{maximum_spanning_tree_based_partitioning}

After the graph $G$ gets constructed, we  partition the graph into two components corresponding to background and foreground by removing minimum weight edge in Maximum Spanning Tree of pixel-segment graph $G$.
\subsubsection{Maximum Spanning Tree Construction}
We first compute the maximum spanning tree of a graph ($MaxST$) which represents the pre-existing connections between the nodes (pixel-segments) in the graph $G$. Hence it has $N+2$ nodes and $N-1$ edges. 
\subsubsection{Maximum Spanning Tree Partitioning}
After getting $MaxST$ from the above part, we remove the lowest weight edge in the tree connecting ‘background’ and ‘foreground’. Thus, we get two clusters, a cluster having ‘background’ node is named as 'background' cluster and the other is named as the ‘foreground’ cluster. In this way, we segment the image into ‘foreground’ and ‘background’. 

We remove the lowest weight edge because it denotes the least similarity between the two nodes. We select the  maximum spanning tree because this represents the maximum similarity tree of segments. As discussed in Section \ref{sec:theoretical_analysis}, removing minimum edge joining background and foreground would optimally partition the graph into two partitions.

\begin{algorithm}[ht] 
\algsetup{linenosize=\normalsize} 
\algsetup{linenodelimiter=)} 
\caption{Iterative and Interactive Image Segmentation (PSSI-MaxST).}
\label{iterative_interactive_image_segmentation}

\begin{tabular}{rl}
    \textbf{Input:} & Image (\( I \)) \\
    \textbf{Output:} & Foreground Mask (\( FM \))
\end{tabular}
\vspace{1mm} 
\hrule 

\begin{algorithmic}[1]
    \STATE Compute Mean-Shift segments on Image \( I \);
    \STATE Compute pixel-segments Graph \( G \) as given in Section \ref{graph-construction} where edge weights are assigned using our proposed Pixel Segment Similarity Index (PSSI) similarity measure;
    \STATE Take User Inputs or Scribbles as given in Section \ref{getting_user_validated_pixel_segments};
    \label{returning_point_in_proposed_algorithm}
    \STATE Connect the user validated pixel-segments to terminal nodes.
    \STATE Construct Maximum Spanning Tree (MaxST) from pixel-segment graph $G$.
    \STATE Partition the MaxST by removing the lowest weight edge to get background ($B$) and foreground ($F$) clusters;
    \STATE Make translucent background with clear foreground and store it in image $MI$; 
    \STATE Display $MI$ to the user.
    \IF{user is not satisfied}
        \STATE \textbf{go to} \ref{returning_point_in_proposed_algorithm}
    \ENDIF 
    \STATE Generate a foreground mask $FM$ using $F$
    
    \STATE \textbf{Return} \( FM \);
    \\
\end{algorithmic}

\end{algorithm}

\subsection{Theoretical Analysis and Energy Formulation}
\label{sec:theoretical_analysis}
Let:
\begin{itemize}
\item $N$ refers to the number of pixel-segments.
    \item $G = (V, E)$, where $V$ denotes vertices ($N$ pixel-segments + 2 terminal nodes, and $E$ represents edges between neighboring pixel-segments)
  \item \(w_{ij} = \mathrm{PSSI}(i,j)\) for \((i,j)\in E\);
  \item \(x_i\in\{0,1\}\) be the binary label of node \(i\) (0=background, 1=foreground);
  \item \(V_t=\{\text{background},\,\text{foreground}\}\) be the two terminal nodes;
  \item \(W_{\max} = \max_{(i,j)\in E} w_{ij}\).
\end{itemize}
Define the total energy
\begin{equation}
E(x)
\;=\;
\underbrace{
\sum_{(i,j)\in E} \bigl|x_i - x_j\bigr|
\;\Bigl(1 - \tfrac{w_{ij}}{W_{\max}}\Bigr)
}_{E_{\mathrm{boundary}}}
\;+\;
\underbrace{
\sum_{i\in V_t} \Phi(x_i)
}_{E_{\mathrm{terminals}}},
\label{eq:total_energy}
\end{equation}
where
\[
\Phi(x_i)
=
\begin{cases}
0, & \text{if } x_i \text{ matches the user‐provided label,}\\
+\infty, & \text{otherwise.}
\end{cases}
\]
Minimizing \(E(x)\) enforces (i) high penalty for cutting edges with large PSSI, and (ii) exact agreement with user scribbles.

\subsubsection{Equivalence to MST‐Cut}
For any cut \(C=\{(i,j)\mid x_i\neq x_j\}\),
\[
E_{\mathrm{boundary}}
=\sum_{(i,j)\in C}\Bigl(1 - \tfrac{w_{ij}}{W_{\max}}\Bigr)
=\;|C|\;-\;\tfrac{1}{W_{\max}}\sum_{(i,j)\in C}w_{ij}.
\]
Thus, among cuts of fixed cardinality, minimizing \(E_{\mathrm{boundary}}\) \(\Leftrightarrow\) maximizing \(\sum_{(i,j)\in C}w_{ij}\).  In the two‐terminal case, it is proven (\cite{zahn1971graph}, \cite{he2008new}) that the maximum‐sum cut is obtained by deleting the \emph{minimum‐weight} edge on the unique path between the terminals in the \emph{Maximum Spanning Tree} (MaxST).  Thus, our approach (build MaxST, remove its smallest edge) \emph{exactly} solves
\[
\min_{x} E(x)
\quad
\]
s.t. each label set is connected and respects scribbles.

\subsubsection{Uniqueness and Stability}
Since each PSSI weight \(w_{ij}>0\) is a continuous (smooth) function of the underlying histograms, generic inputs yield distinct weights, hence a unique MaxST up to ties.  If ties occur (equal PSSI), any choice among equal‐weight edges remains optimal, as they correspond to identical energy gaps.

\subsubsection{Algorithmic Complexity}
\begin{itemize}
  \item \textbf{MaxST construction:} Kruskal’s algorithm in \(O(|E|\log|E|)\).  On our superpixel graph, \(|E|=O(N)\), so \(O(N\log N)\).
  \item \textbf{Cut extraction:} Scanning edges on the MaxST (\(N-1\) edges) to find the global minimum takes \(O(N)\).
\end{itemize}
Therefore, the overall MST‐cut step runs in \(O(N\log N)\), making it highly efficient for interactive use.

\begin{table}[htb]
    \caption{Summary of evaluation metrics for segmentation performance, including Jaccard Index (IoU), precision (P), recall (R), $F_{1}$ and $F_{\beta}$ scores, and mean error rate.}
    \label{tab:evaluation-metrics}
    \centering
    \renewcommand{\arraystretch}{1.4}
    \resizebox{0.4\textwidth}{!}{%
    \begin{tabular}{@{}ll@{}}
        \toprule
        \textbf{Evaluation Metric} & \textbf{Formula} \\
        \midrule
        Jaccard Index ($IoU$) 
            & $\displaystyle \frac{TP}{TP + FP + FN}$ \\[4pt]
        Precision ($P$)
            & $\displaystyle \frac{TP}{TP + FP}$ \\[4pt]
        Recall ($R$)
            & $\displaystyle \frac{TP}{TP + FN}$ \\[4pt]
        $F_{1}$ Score 
            & $\displaystyle \frac{2 \, P \, R}{P + R}$ \\[4pt]
        $F_{\beta}$ Score ($\beta^{2}= 0.3$) 
            & $\displaystyle \frac{(1+\beta^{2})\,P\,R}{\beta^{2}\,P + R}$ \\[4pt]
        Mean Error ($ME$)
            & $\displaystyle \frac{FP + FN}{TP + TN + FP + FN}$ \\
        \bottomrule
    \end{tabular}%
    }
\end{table}

\section{Experimental Results and Analysis}
\label{experimental_analysis}

\subsection{Dataset Description and Hardware Configuration}
\label{sec:dataset_hardware}
We evaluate our method on two benchmark datasets. First, Images250\cite{qu2022interactive} comprises 250 images randomly selected from the Berkeley Segmentation Dataset (BSD)\cite{martin2001database} and the MSRA dataset \cite{4270072}. The authors of Images250 provide optimal user scribbles for interactive graph‑based methods including AMOE, OneCut, and SSNCut \cite{qu2022interactive}. This collection covers three image categories: (i) simple scenes with single, low‑texture backgrounds; (ii) complex scenes with highly textured backgrounds; and (iii) challenging cases where foreground and background have very similar colors. Image content spans animals, buildings, people, fruit, airplanes, flowers, cars, and plants \cite{qu2022interactive}. 

Second, we utilize the GrabCut dataset \cite{rother2004grabcut}, for which optimal scribbles were obtained from Perret et al. (2015) \cite{perret2015evaluation}. Both datasets provide binary ground‑truth masks for quantitative evaluation. 

All baseline methods (AMOE, OneCut, and SSN–Cut) were executed via MATLAB Online, while our proposed PSSI–MaxST and BHA–MaxST variants were implemented in Python~3.10.7 on a computer equipped with an 11\textsuperscript{th} Gen Intel\textregistered{} Core\texttrademark{} i5-1135G7 @ 2.40\,GHz processor, 8\,GB RAM, and a 64-bit operating system. 

\subsection{Evaluation measures}
The evaluation metrics utilized in this work is listed in Table \ref{tab:evaluation-metrics}. For $F_\beta$ we use $\beta^2 = 3$ to weight precision slightly more than recall similar to Qu et al. (2022) \cite{qu2022interactive}.

\begin{figure}[!t]
    \centering
    \includegraphics[width=\linewidth]{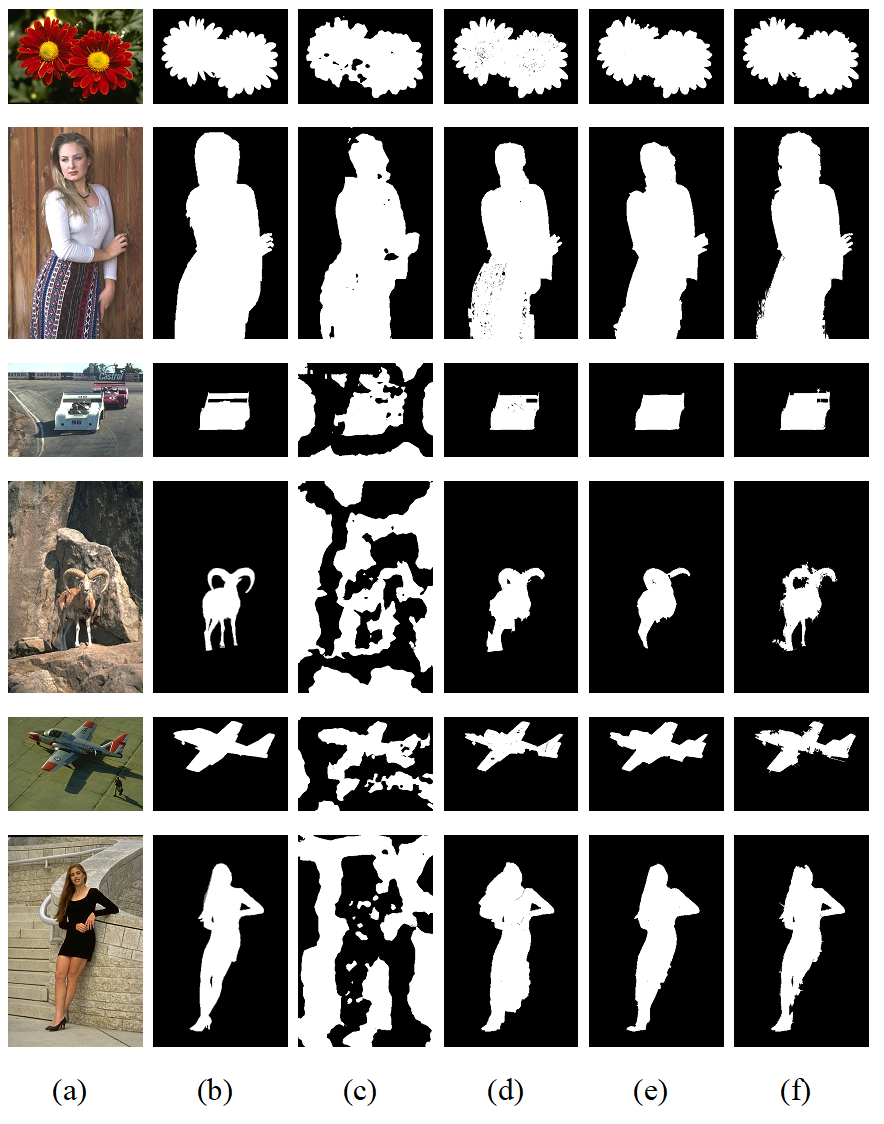}
    \caption{Segmentation results on the GrabCut dataset. (a) Original image, (b) Ground truth, (c) SSNCut output, (d) OneCut output, (e) AMOE output, and (f) Our (PSSI) method.}
    \label{fig:grabcut_results}
\end{figure}

\begin{figure}[!t]
    \centering
    \includegraphics[width=\linewidth]{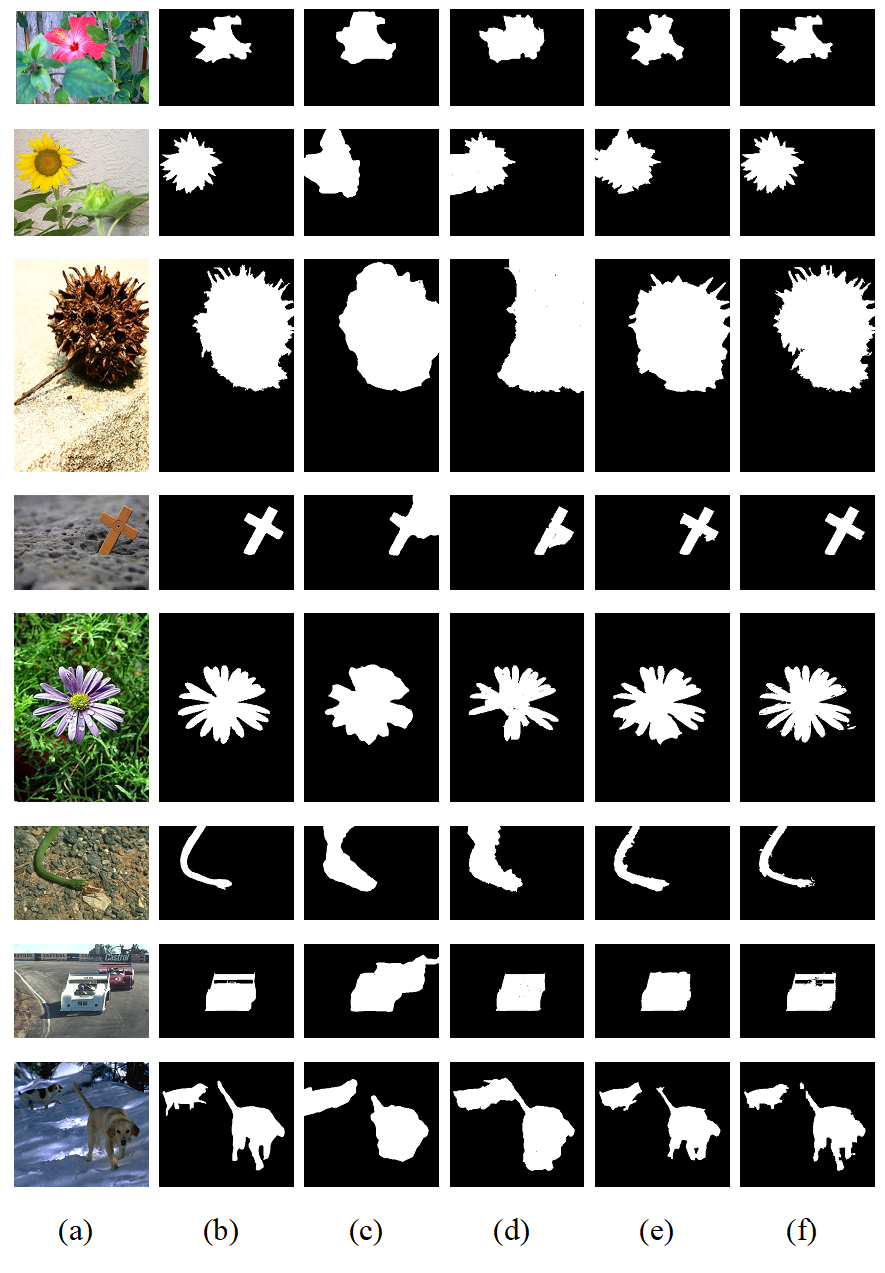}
    \caption{Segmentation results on the Images250 dataset using optimal scribbles for each method. (a) Original image, (b) Ground truth, (c) SSNCut output, (d) OneCut output, (e) AMOE output, and (f) Our (PSSI) method.}
    \label{fig:images250_results}
\end{figure}

\begin{figure}[!t]
    \centering
    \includegraphics[width=\linewidth]{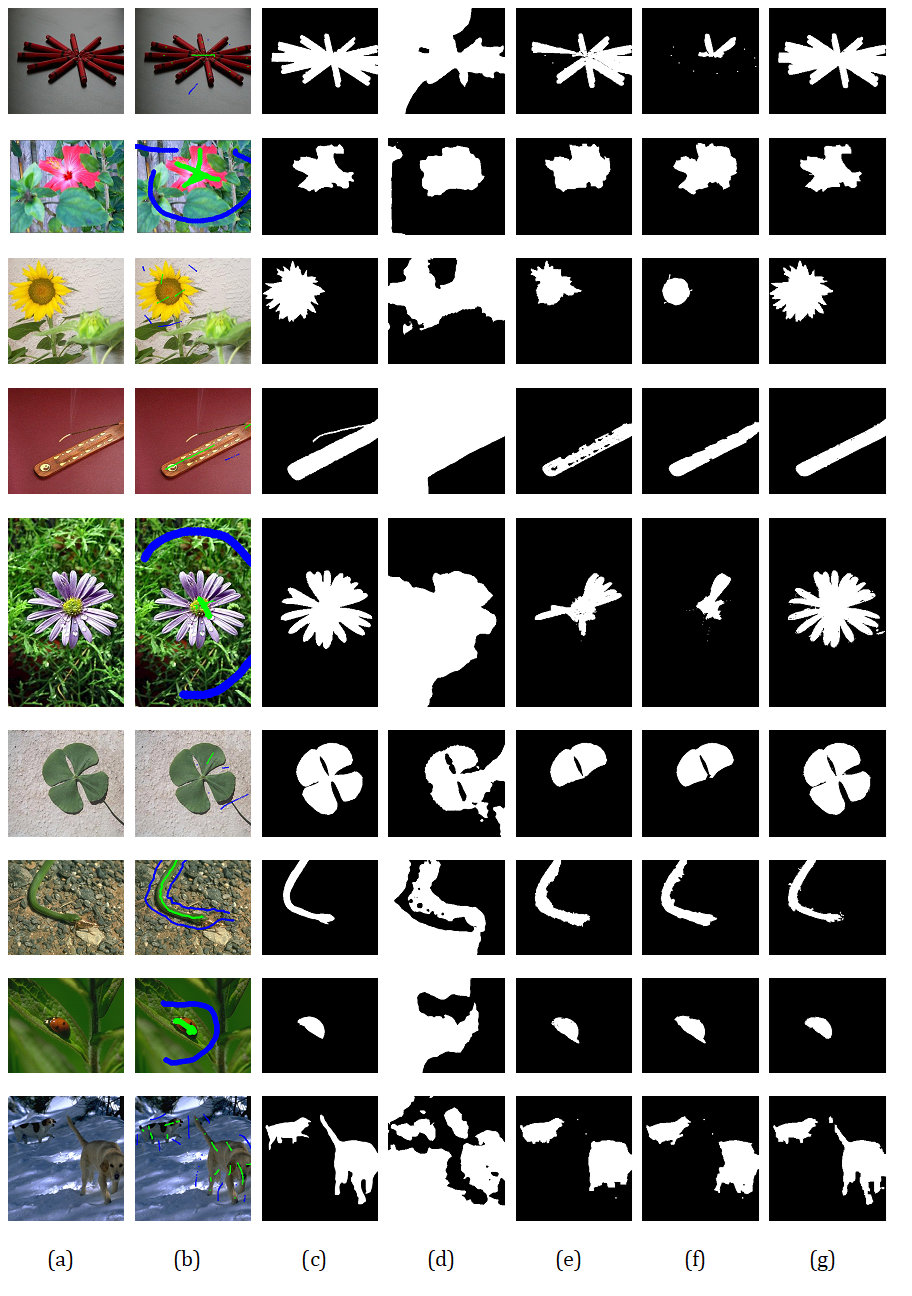}
    \caption{Segmentation results with scribbles from our method. (a) Original image, (b) Markers, (c) Ground truth, (d) SSNCut output, (e) OneCut output, (f) AMOE output, and (g) our method.}
    \label{fig:scribble_results}
\end{figure}

\begin{figure}[!t]
    \centering
    \includegraphics[width=\linewidth]{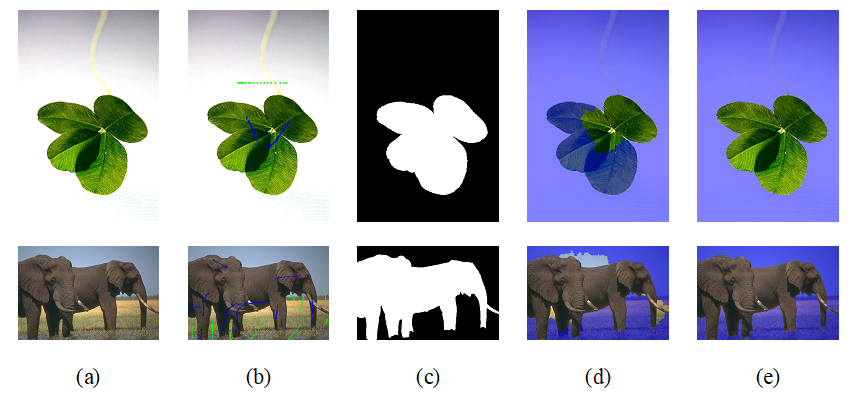}
    \caption{Segmentation results with scribbles from our method. (a) Original image, (b) Markers, (c) Ground truth, (d) PSSI-GraphCut output, (e) PSSI-MaxST output.}
    \label{fig:graphcut_comparison}
\end{figure}

\begin{table}[!t]
\centering
\scriptsize
\caption{Jaccard Index on the \texttt{images250} dataset. Higher is better for Avg/Min/Max ($\uparrow$), lower is better for Std. Dev. ($\downarrow$). Both BHA and PSSI outperform baselines.}
\label{tab:jaccard_images250}
\begin{tabular}{lcccc}
\toprule
Method         & Avg ($\uparrow$)    & Std. Dev. ($\downarrow$) & Min ($\uparrow$)     & Max ($\uparrow$)     \\
\midrule
\rowcolor{green!15}

Our (PSSI-MaxST)     & \textbf{0.9302}     & \textbf{0.0558}          & \textbf{0.7295}      & \textbf{0.9961}      \\
\rowcolor{green!10}
Our (BHA-MaxST)      & \textbf{0.9263}     & \textbf{0.0567}                & \textbf{0.6998}          & \textbf{0.9959}            \\ 
AMOE           & 0.9243              & 0.0593                   & 0.6619              & 0.9923               \\
SSN‑Cut        & 0.7908              & 0.2556                   & 0.0049               & 0.9869               \\
OneCut         & 0.8623              & 0.1110                   & 0.3649               & 0.9944               \\
\bottomrule
\end{tabular}
\end{table}

\begin{table}[!t]
\centering
\scriptsize
\caption{Precision, Recall, F\textsubscript{1}, F\textsubscript{$\beta$} and ME on the \texttt{images250} dataset. 
Both BHA and PSSI similarity measures outperform all baselines.}
\label{tab:precision_images250}
\begin{tabular}{lccccc}
\toprule
Method            & Precision ($\uparrow$) & Recall ($\uparrow$) & F\textsubscript{1} ($\uparrow$) & F\textsubscript{$\beta$} ($\uparrow$) & ME ($\downarrow$) \\
\midrule
\rowcolor{green!15}
Our (PSSI-MaxST)        & \textbf{0.9682}        & \textbf{0.9592}     & \textbf{0.9631}                 & \textbf{0.9661}                       & \textbf{0.0112}   \\
\rowcolor{green!10}

Our (BHA-MaxST)         & 0.9645 & \textbf{0.9586} & \textbf{0.9608} & 0.9631 & 0.0122   \\

AMOE              & \textbf{0.9672}                 & 0.9535              & 0.9596                          & \textbf{0.9640 }                               & \textbf{0.0121}         \\
SSN‑Cut           & 0.8229                 & 0.9342              & 0.8505                          & 0.8462                                & 0.0668            \\
OneCut            & 0.9228                 & 0.9299              & 0.9219                          & 0.9244                                & 0.0268            \\
\bottomrule
\end{tabular}
\end{table}
 \begin{table}[!t]
\centering
\scriptsize
\caption{Jaccard Index on the \texttt{grabcut} dataset. Higher is better for Avg/Min/Max ($\uparrow$), lower is better for Std. Dev. ($\downarrow$).}
\label{tab:jaccard_grabcut}
\begin{tabular}{lcccc}
\toprule
Method       & Avg ($\uparrow$)      & Std. Dev. ($\downarrow$) & Min ($\uparrow$)      & Max ($\uparrow$)      \\
\midrule
\rowcolor{green!15}
Our (BHA-MaxST)      & \textbf{0.9431}     & \textbf{0.0525}                & 0.7295            & 0.9928 \\
\rowcolor{green!10}
Our (PSSI-MaxST)   & \textbf{0.9237}       & \textbf{0.0572}          & \textbf{0.7704}       & \textbf{1.0000}       \\
AMOE         & 0.9026                & 0.0652                   & \textbf{0.7465}                & 0.9891                \\
SSN‑Cut      & 0.5613                & 0.3708                   & 0.0013                & 0.9767                \\
OneCut       & 0.8918                & 0.0805                   & 0.7321                & \textbf{0.9936  }              \\
\bottomrule
\end{tabular}
\end{table}

\begin{table}[!t]
\centering
\scriptsize
\caption{Precision, Recall, F\textsubscript{1}, F\textsubscript{$\beta$} and ME on the \texttt{grabcut} dataset. Higher is better for all but ME ($\downarrow$).}
\label{tab:precision_grabcut}
\begin{tabular}{lccccc}
\toprule
Method       & Precision ($\uparrow$) & Recall ($\uparrow$) & F\textsubscript{1} ($\uparrow$) & F\textsubscript{$\beta$} ($\uparrow$) & ME ($\downarrow$) \\
\midrule

\rowcolor{green!15}
Our (BHA-MaxST)         & \textbf{0.9756}        & \textbf{0.9688}     & \textbf{0.9700}                 & \textbf{0.9740}                       & \textbf{0.0104}   \\

\rowcolor{green!10}
Our (PSSI-MaxST)   & \textbf{0.9682}        & \textbf{0.9522}     & \textbf{0.9594}                 & \textbf{0.9644}                       & \textbf{0.0144}   \\
AMOE         & 0.9601                 & 0.9369              & 0.9475                          & 0.9546                                & 0.0179            \\
SSN‑Cut      & 0.5777                 & 0.8233              & 0.6385                          & 0.6204                                & 0.2322            \\
OneCut       & 0.9534                 & 0.9343              & 0.9408                          & 0.9489                                & 0.0209            \\
\bottomrule
\end{tabular}
\end{table}

 \begin{table}[!t]
\centering
\scriptsize
\caption{Jaccard Index under exchange of markers on the \texttt{images250} dataset.}
\label{tab:jaccard_exchange_of_markers_images250}
\begin{tabular}{llcccc}
\toprule
Method        & Markers     & Avg ($\uparrow$)    & Std. Dev. ($\downarrow$) & Min ($\uparrow$)    & Max ($\uparrow$)    \\
\midrule
\multirow{3}{*}{Our (PSSI-MaxST)}
              & AMOE        & 0.9157              & 0.0678                   & 0.6633              & 0.9961              \\
              & SSN‑Cut     & 0.8769              & 0.1078                   & 0.4156              & 0.9961              \\
              & OneCut      & 0.8781              & 0.1003                   & 0.5561              & 0.9961              \\
\midrule
\multirow{3}{*}{AMOE}
              & Our         & 0.8948              & 0.1205                   & 0.1038              & 0.9923              \\
              & SSN‑Cut     & 0.8353              & 0.1574                   & 0.2109              & 0.9923              \\
              & OneCut      & 0.8659              & 0.1110                   & 0.3525              & 0.9915              \\
\midrule
\multirow{3}{*}{SSN‑Cut}
              & Our         & 0.4963              & 0.3598                   & 0.0045              & 0.9861              \\
              & AMOE        & 0.5208              & 0.3449                   & 0.0069              & 0.9869              \\
              & OneCut      & 0.5999              & 0.3301                   & 0.0008              & 0.9841              \\
\midrule
\multirow{3}{*}{OneCut}
              & Our         & 0.8698              & 0.1553                   & 0.0452              & 0.9947              \\
              & AMOE        & 0.8984              & 0.0845                   & 0.3500              & 0.9950              \\
              & SSN‑Cut     & 0.8856              & 0.1205                   & 0.2324              & 0.9975              \\
\bottomrule
\end{tabular}
\end{table}

\definecolor{OurColor}{HTML}{D0E1F9}     
\definecolor{AMOEColor}{HTML}{F9D0D0}    
\definecolor{SSNCutColor}{HTML}{D0F9D2}  
\definecolor{OneCutColor}{HTML}{F9E5D0}  

\begin{table}[!t]
\centering
\scriptsize
\caption{Precision, Recall, F\textsubscript{1}, F\textsubscript{$\beta$} and ME under exchange of markers on the \texttt{images250} dataset.}
\label{tab:precision_exchange_of_markers_images250}
\resizebox{0.49\textwidth}{!}{%
\begin{tabular}{llccccc}
\toprule
Method        & Markers     & Precision ($\uparrow$) & Recall ($\uparrow$) & F\textsubscript{1} ($\uparrow$) & F\textsubscript{$\beta$} ($\uparrow$) & ME ($\downarrow$) \\
\midrule
\multirow{3}{*}{Our (PSSI-MaxST)}
              & AMOE        & 0.9578               & 0.9532              & 0.9546                          & 0.9657                          & 0.0140            \\
              & SSN‑Cut     & 0.9540                 & 0.9137              & 0.9306                          & 0.9444                          & 0.0200            \\
              & OneCut      & 0.9215                 & 0.9488              & 0.9318                          & 0.9277                          & 0.0223            \\
\midrule
\multirow{3}{*}{AMOE}
              & Our         & 0.9665                & 0.9239              & 0.9386                          & 0.9563                          & 0.0187            \\
              & SSN‑Cut     & 0.9551                 & 0.8688              & 0.9007                          & 0.9336                          & 0.0278            \\
              & OneCut      & 0.9203                 & 0.9368              & 0.9239                          & 0.9241                          & 0.0256            \\
\midrule
\multirow{3}{*}{SSN‑Cut}
              & Our       & 0.5154                 & 0.8115              & 0.5829                          & 0.5268                          & 0.2665            \\
              & AMOE        & 0.5388                 & 0.8405              & 0.6122                          & 0.5875                          & 0.2319            \\
              & OneCut      & 0.6117                 & 0.9258              & 0.6875                          & 0.6637                          & 0.1815            \\
\midrule
\multirow{3}{*}{OneCut}
              & Our         & 0.9653                 & 0.8981              & 0.9186                          & 0.9489                          & 0.0265            \\
              & AMOE        & 0.9665                 & 0.9285              & 0.9441                          & 0.9575                          & 0.0182            \\
              & SSN‑Cut     & 0.9544                 & 0.8482              & 0.8856                          & 0.9276                          & 0.0350            \\
\bottomrule
\end{tabular}
}
\end{table}

 \begin{table*}[!t]
\centering
\scriptsize
\caption{Performance comparison of SLIC and MeanShift as low-level segmentation methods. All metrics favor higher values, except for ME, where lower values indicate better results.}
\label{tab:comparison_slic_meanshift}
\resizebox{\textwidth}{!}{%
\begin{tabular}{lllcccccc}
\toprule
Dataset & Similarity Measure & LL-Seg. & Jaccard ($\uparrow$) & Precision ($\uparrow$) & Recall ($\uparrow$) & F\textsubscript{1} ($\uparrow$) & F\textsubscript{$\beta$} ($\uparrow$) & ME ($\downarrow$) \\
\midrule

\multirow{4}{*}{Images250}
  & \multirow{2}{*}{PSSI}
     & Mean-Shift & \textbf{0.9302} & \textbf{0.9674} & \textbf{0.9596} & \textbf{0.9629} & \textbf{0.9656} & \textbf{0.0113} \\
  &  & SLIC & 0.9051 & 0.9581 & 0.9422 & 0.9488 & 0.9544 & 0.0160 \\
  \cmidrule(lr){2-9}
  & \multirow{2}{*}{BHA}
     & Mean-Shift & \textbf{0.9292} & \textbf{0.9691} & \textbf{0.9572} & \textbf{0.9623} & \textbf{0.9663} & \textbf{0.0116} \\
  &  & SLIC & 0.9047 & 0.9518 & 0.9474 & 0.9486 & 0.9508 & 0.0155 \\

\midrule

\multirow{4}{*}{GrabCut}
  & \multirow{2}{*}{PSSI}
     & Mean-Shift & \textbf{0.9237} & \textbf{0.9682} & \textbf{0.9522} & \textbf{0.9594} & \textbf{0.9644} & \textbf{0.0144} \\
  &  & SLIC & 0.9034 & 0.9612 & 0.9373 & 0.9481 & 0.9556 & 0.0183 \\
  \cmidrule(lr){2-9}
  & \multirow{2}{*}{BHA}
     & Mean-Shift & \textbf{0.9431} & \textbf{0.9756} & \textbf{0.9688} & \textbf{0.9700} & \textbf{0.9740} & \textbf{0.0104} \\
  &  & SLIC & 0.9240 & 0.9604 & 0.9600 & 0.9597 & 0.9631 & 0.0139 \\
\bottomrule
\end{tabular}%
}
\end{table*}

\begin{table*}[!t]
\centering
\scriptsize
\caption{Performance comparison of MaxST and GraphCut for partitioning. All metrics favor higher values, except for ME, where lower values indicate better results.}
\label{tab:comparison_maxst_graphcut}
\resizebox{\textwidth}{!}{%
\begin{tabular}{lllcccccc}
\toprule
Dataset     & Similarity Measure & LL‑Seg.      & Jaccard ($\uparrow$) & Precision ($\uparrow$) & Recall ($\uparrow$) & F\textsubscript{1} ($\uparrow$) & F\textsubscript{$\beta$} ($\uparrow$) & ME ($\downarrow$) \\
\midrule

\multirow{4}{*}{Images250}
  & \multirow{2}{*}{PSSI}
     & MaxST      & \textbf{0.9302} & \textbf{0.9674} & \textbf{0.9596} & \textbf{0.9629} & \textbf{0.9656} & \textbf{0.0113} \\
  & 
     & GraphCut   & 0.7439          & 0.9277          & 0.8030          & 0.8290          & 0.0531          & 0.0531          \\
  \cmidrule(lr){2-9}
  & \multirow{2}{*}{BHA}
     & MaxST      & \textbf{0.9292} & \textbf{0.9691} & \textbf{0.9572} & \textbf{0.9623} & \textbf{0.9663} & \textbf{0.0116} \\
  &
     & GraphCut   & 0.9020          & 0.9739          & 0.9243          & 0.9466          & 0.9620          & 0.0160          \\

\midrule

\multirow{4}{*}{GrabCut}
  & \multirow{2}{*}{PSSI}
     & Mean‑Shift & \textbf{0.9237} & \textbf{0.9682} & \textbf{0.9522} & \textbf{0.9594} & \textbf{0.9644} & \textbf{0.0144} \\
  &
     & GraphCut   & 0.7792          & 0.8650          & 0.8275          & 0.8281          & 0.8561          & 0.0587          \\
  \cmidrule(lr){2-9}
  & \multirow{2}{*}{BHA}
     & Mean‑Shift & \textbf{0.9431} & \textbf{0.9756} & \textbf{0.9688} & \textbf{0.9700} & \textbf{0.9740} & \textbf{0.0104} \\
  &
     & GraphCut   & 0.9326          & 0.9764          & 0.9537          & 0.9644          & 0.9711          & 0.0130          \\
\bottomrule
\end{tabular}%
}
\end{table*}

\begin{table*}[!t]
\centering
\scriptsize
\caption{Comparison of segmentation performance using PSSI and Bhattacharyya (BHA) similarity measures, with Mean‑Shift superpixels and MaxST partitioning. All metrics favor higher values except ME.}
\label{tab:bha_PSSI_comparison}
\resizebox{\textwidth}{!}{%
\begin{tabular}{@{}l l c c c c c c@{}}
    \toprule
    Dataset & Similarity Measure & Jaccard ($\uparrow$) & Precision ($\uparrow$) & Recall ($\uparrow$) & F\textsubscript{1} ($\uparrow$) & F\textsubscript{$\beta$} ($\uparrow$) & ME ($\downarrow$) \\
    \midrule
    \multirow{2}{*}{Images250}
      & PSSI & 0.9302 & 0.9674 & 0.9596 & 0.9629 & 0.9656 & 0.0113 \\
      & BHA  & 0.9292 & 0.9691 & 0.9572 & 0.9623 & 0.9663 & 0.0116 \\
    \midrule
    \multirow{2}{*}{GrabCut}
      & PSSI & 0.9237 & 0.9682 & 0.9522 & 0.9594 & 0.9644 & 0.0144 \\
      & BHA  & 0.9431 & 0.9756 & 0.9688 & 0.9700 & 0.9740 & 0.0104 \\
    \bottomrule
\end{tabular}%
}
\end{table*}

\subsection{Evaluation on the Image250 and GrabCut Datasets}

To assess the effectiveness of our proposed methods, we conducted a comprehensive evaluation on two standard benchmarks: Image250 \cite{qu2022interactive} and GrabCut \cite{rother2004grabcut}. We compare five approaches i.e., our PSSI-based variant (PSSI), our Bhattacharyya‐based variant (BHA), AMOE \cite{qu2022interactive}, SSNCut \cite{chew2015semi}, and OneCut \cite{tang2013grabcut}, using evaluation metrics listed in Table \ref{tab:evaluation-metrics}. Tables~\ref{tab:jaccard_images250} and \ref{tab:precision_images250} summarize the results on the Image250 dataset, while Tables~\ref{tab:jaccard_grabcut} and \ref{tab:precision_grabcut} report the corresponding scores on GrabCut.

On both the Image250 and GrabCut benchmarks, our two variants (PSSI and BHA) consistently outperform all baselines in terms of overlap and boundary-based metrics. On Image250 (Tables~\ref{tab:jaccard_images250} and~\ref{tab:precision_images250}), PSSI attains the highest mean Jaccard index (0.9306) with the lowest standard deviation (0.0553), and BHA follows closely (0.9263~$\pm$~0.0567). Our BHA variant exceeds AMOE (0.9243~$\pm$~0.0593) by more than 0.2~points while our PSSI variant exceeds by more than 0.6 points, and markedly outperform SSN\textendash Cut (0.7908~$\pm$~0.2556) and OneCut (0.8623~$\pm$~0.1110). In boundary measures, PSSI-MaxST and BHA-MaxST achieve $F_1$ scores of 0.9631 and 0.9608, respectively. PSSI-MaxST achieves the lowest mean error (ME) $0.0112$, improving upon AMOE’s $F_1 = 0.9596$ and $\mathrm{ME} = 0.0121$, while BHA variant gets mean error of $0.0122$.

Similarly, on GrabCut (Tables~\ref{tab:jaccard_grabcut} and~\ref{tab:precision_grabcut}), BHA leads with a mean Jaccard of 0.9431 ($\sigma = 0.0525$) and PSSI follows at 0.9237 ($\sigma = 0.0572$), compared to AMOE’s 0.9026 ($\sigma = 0.0652$). BHA also attains the highest $F_1$ score (0.9700) and lowest mean error (0.0104), while PSSI records 0.9594 and 0.0144, respectively. Both variants not only boost average overlap by over 1.0~point relative to AMOE, but also exhibit higher minimum scores and tighter distributions, evidencing both improved accuracy and robustness.

Visual segmentation results on Image250 and GrabCut are shown in Figures~\ref{fig:images250_results} and~\ref{fig:grabcut_results}, respectively. Qualitative comparison demonstrates that our method with PSSI yields more accurate and coherent object boundaries than competing approaches. In particular, our method better preserves fine details (e.g., thin structures and complex contours) and more effectively suppresses background artifacts. These improvements are consistent across diverse scenes showing the effectiveness of our approach.

\subsection{Exchange of Seed‐Point Markers}
\label{sec:exchange_seed_point}

To provide a fair and objective comparison, we adopt the marker configurations provided by Qu et al. (2020) \cite{qu2022interactive} for AMOE, OneCut, and SSN‐Cut on the Image250 dataset. We manually annotated for optimal markers for our method.  This was done to ensure the generalizability and robustness of various methods to various markers. Quantitative results are shown in Table \ref{tab:jaccard_exchange_of_markers_images250}) Table \ref{tab:precision_exchange_of_markers_images250}.

On the Image250 dataset, even when using seed‑point markers from competing methods, our method maintains superior overlap performance (Table~\ref{tab:jaccard_exchange_of_markers_images250}). When supplied with AMOE’s markers, PSSI-MaxST achieves a mean Jaccard of 0.9157 ($\sigma = 0.0678$), outperforming AMOE on our seeds (0.8948~$\pm$~0.1205) and OneCut on its own markers (0.8659~$\pm$~0.1110). With SSN‑Cut markers, PSSI-MaxST still attains 0.8769~$\pm$~0.1078 significantly above SSN‑Cut’s best of 0.5999~$\pm$~0.3301. Likewise, using OneCut markers yields 0.8781~$\pm$~0.1003, marginally exceeding OneCut’s 0.8856~$\pm$~0.1205 despite marker exchange. These results confirm that our similarity measure and graph‑cut optimization are more robust to initialization than the baselines.

Boundary‑based metrics (Table~\ref{tab:precision_exchange_of_markers_images250}) show a similar trend. PSSI-MaxST with AMOE markers achieves $F_1 = 0.9546$ and $\mathrm{ME} = 0.0140$, outperforming AMOE with our markers ($F_1 = 0.9386$, $\mathrm{ME} = 0.0187$) and OneCut with its own markers ($F_1 = 0.9239$, $\mathrm{ME} = 0.0256$). Even under the most challenging SSN‑Cut initialization, PSSI-MaxST records $F_1 = 0.9306$ and $\mathrm{ME} = 0.0200$, compared to SSN‑Cut’s best $F_1 = 0.6875$. Overall, these ``marker‑exchange'' experiments demonstrate that PSSI delivers consistently higher accuracy and lower error than all baselines, regardless of the seed‑point configuration.

\subsection{Analysing the Shortcomings of Current Methods}

Figure~\ref{fig:scribble_results} presents segmentation outputs of competing methods (SSN‑Cut, OneCut, and AMOE) when initialized with the scribble annotations from our approach. It is evident that SSN‑Cut struggles on images with weak boundary cues, frequently leaking into the background. OneCut often misclassifies foreground regions as background, particularly in examples 5 and 6, due to its $L_1$ distance similarity and ignoring of pixel-distance. AMOE similarly fails to capture complete object extents in images 1, 2, 5, and 6, as its pixel‐level graph construction is overly sensitive to local intensity variations. By contrast, our method consistently delineates the foreground accurately across all challenging cases, demonstrating robustness to weak edges and complex appearance variations.

\subsection{Comparing SLIC and MeanShift for Low‐Level Segmentation}

Table~\ref{tab:comparison_slic_meanshift} compares the performance of our two variants (PSSI and BHA) when using SLIC versus MeanShift for superpixel generation on both the Image250 and GrabCut datasets. Across all metrics, MeanShift consistently outperforms SLIC. For example, on Image250 with PSSI, MeanShift achieves a Jaccard of 0.9302 versus 0.9051 for SLIC, and reduces ME from 0.0160 to 0.0113. Similarly, on GrabCut with BHA, MeanShift improves the Jaccard from 0.9240 to 0.9431 and lowers ME from 0.0139 to 0.0104. We attribute this gain to MeanShift’s ability to better preserve object shapes and adhere to natural image boundaries, which yields more coherent superpixels and, consequently, more accurate graph‐cut segmentation.

\subsection{Comparing GraphCut with MaxST for partitioning}

Figure~\ref{fig:graphcut_comparison} illustrates example segmentations produced by our MaxST‐based algorithm versus the GraphCut approach. Quantitative results in Table~\ref{tab:comparison_maxst_graphcut} confirm that MaxST partitioning consistently outperforms GraphCut across both datasets and similarity measures. On Image250, PSSI + MaxST achieves a Jaccard index of 0.9302 compared to 0.7439 for PSSI + GraphCut, with corresponding $F_1$ scores of 0.9629 versus 0.8290 and mean errors (ME) of 0.0113 versus 0.0531. Similarly, BHA + MaxST records 0.9292 Jaccard against 0.9020 for PSSI + GraphCut, improving $F_1$ from 0.9466 to 0.9623 and reducing ME from 0.0160 to 0.0116. On GrabCut, PSSI + MaxST yields 0.9237 Jaccard (PSSI + GraphCut: 0.7792) and $F_1=0.9594$ (PSSI + GraphCut: 0.8281), while BHA + MaxST rises to 0.9431 Jaccard and $F_1=0.9700$, surpassing PSSI + GraphCut’s 0.9326 and 0.9644, respectively. These improvements demonstrate that MaxST partitioning on the segment graph produces more accurate and stable foreground–background cuts than the GraphCut, due to its global optimization over segment connectivity and energy minimization achieved discussed in Section \ref{sec:theoretical_analysis}.

\subsection{Comparison of the proposed similarity measure with Bhattacharyya Measure}
\label{subsec:comparison_with_previous_measure}
As discussed in the Section \ref{proposed_similarity_measure}, the time complexity of Bhattacharyya Measure is cubic in number of bins making it slow. In contrast, the proposed similarity measure is linear in number of bins making it computationally efficient for the task.

Table~\ref{tab:bha_PSSI_comparison} reports segmentation performance on the Image250 and GrabCut datasets using MeanShift superpixels and MaxST partitioning. Both PSSI and BHA achieve comparable accuracy: on Image250, PSSI attains a Jaccard index of 0.9302 versus 0.9292 for BHA, with F$_1$ scores of 0.9629 and 0.9623, respectively. On GrabCut, BHA slightly outperforms PSSI in overlap (0.9431 vs.\ 0.9237) and boundary metrics ($F_1=0.9700$ vs.\ 0.9594). PSSI delivers these results with linear complexity in number of histogram bins, preserving segmentation quality while significantly reducing computational complexity compared to the Bhattacharyya Measure.

\subsection{Scribble Amount Comparison}
\label{subsec:scribble_amount}

As discussed in Section~\ref{sec:exchange_seed_point}, we adopt the marker configurations provided by Qu et al.~\cite{qu2022interactive} for AMOE, OneCut, and SSN‑Cut on the Image250 dataset. We manually annotated optimal scribble markers for our method. The results in Table~\ref{tab:scribble_amounts} clearly demonstrate that our proposed method requires significantly less user input compared to existing approaches. Specifically, our method achieves accurate segmentation with only \textbf{3.04\%} of image pixels marked as scribbles, which is approximately \textbf{6$\times$ less} than OneCut, \textbf{3$\times$ less} than SSN‑Cut, and \textbf{1.6$\times$ less} than AMOE. This highlights the efficiency of our technique in utilizing minimal supervision for effective segmentation. The reduction in scribble burden directly translates to improved usability and faster annotation, making our method more practical for real-world interactive applications.

\begin{table}[!t]
  \centering
  \scriptsize
  \caption{Comparison of scribble amount (percentage of image pixels) required by each method.}
  \label{tab:scribble_amounts}
  \begin{tabular}{lccc}
    \toprule
    \textbf{Method} & \textbf{Background (\%)} & \textbf{Foreground (\%)} & \textbf{Total (\%)} \\
    \midrule
    \rowcolor{green!15} Ours      & 1.86 & 1.18 & 3.04 \\
    AMOE      & 3.06  & 1.98 & 5.03  \\
    SSN‑Cut   & 6.73  & 1.76 & 8.49  \\
    OneCut    & 10.79 & 6.28 & 17.06 \\
    
    \bottomrule
  \end{tabular}
\end{table}

\subsection{Runtime comparison}
Table~\ref{tab:runtime_comparison} reports the runtime performance of various segmentation methods.
While the differences in execution environments (as discussed in Section \ref{sec:dataset_hardware}) limit a strictly fair comparison, the results suggest that our PSSI–MaxST method offers competitive and potentially superior runtime efficiency. In particular, PSSI–MaxST consistently achieves faster segmentation times than both current graph-based interactive image segmentation methods and our own Bhattacharyya-based variant, indicating the computational advantage of the proposed PSSI similarity measure. AMOE and OneCut exhibit higher runtimes because they model each pixel as a graph node, thereby increasing graph‐construction and path‐finding complexity, whereas SSN–Cut’s elevated execution time stems from its computationally intensive globalized probability of boundary (gPb) calculation for assigning edge-weights.

\begin{table}[!t]
  \caption{Runtime Comparison of Various Methods}
  \label{tab:runtime_comparison}
  \centering
  \begin{tabular}{lcccc}
    \hline
    \textbf{Method}              & \textbf{Time (s)} & \textbf{Std.\ Dev.\ (s)} & \textbf{Min (s)} & \textbf{Max (s)} \\
    \hline
    Our (PSSI–MaxST)     &    2.14      &    1.40             &   0.14       & 9.68        \\
    Our (BHA–MaxST)      &     4.07     &       3.30          &   0.20      & 20.75        \\
    AMOE                 &    7.22      &     2.46            &    2.03     &   14.15       \\
    OneCut               &     8.19      &       4.54          &    0.46     &     31.12    \\
    SSN–Cut              &     89.20     &         1.47        &    87.64     & 91.56         \\
    \hline
  \end{tabular}
\end{table}

\subsection{Analyzing Performance across Number of Bins}

\begin{table}[!t]
  \caption{Performance of PSSI–MaxST vs.\ BHA–MaxST for Varying Number of Histogram Bins}
  \label{tab:bins_comparison}
  \centering
  \resizebox{0.38 \textwidth}{!}{
    \begin{tabular}{@{} c l r r r @{}}
    \toprule
    \textbf{\#Bins} & \textbf{Method}         & \textbf{Jaccard} & \textbf{F$_1$}   & \textbf{ME}      \\ 
    \midrule
    \multirow{2}{*}{4}  & PSSI–MaxST & 0.9063 & 0.9494 & 0.0162 \\
                        & BHA–MaxST  & 0.9169  & 0.9554 & 0.0141 \\
    \addlinespace
    \multirow{2}{*}{8}  & PSSI–MaxST & \textbf{0.9306} & 0.\textbf{9631} & \textbf{0.0113} \\
                        & BHA–MaxST  & 0.9263 & 0.9608 & 0.0122 \\
    \addlinespace
    \multirow{2}{*}{16} & PSSI–MaxST & 0.9286 & 0.9620 & 0.0118 \\
                        & BHA–MaxST  &    0.9236    &   0.9594     &   0.0124     \\
    \addlinespace
    \multirow{2}{*}{32} & PSSI–MaxST &   0.9226     &     0.9587   &  0.0130      \\
                        & BHA–MaxST  &    0.9296    &    0.9626    &  0.0116      \\
    \addlinespace
    \multirow{2}{*}{64} & PSSI–MaxST &     0.9286   &  0.9621      & 0.0118       \\
                        & BHA–MaxST  &   0.9205     &    0.9574    &  0.0133      \\
    \bottomrule
  \end{tabular}
  }
\end{table}
It can be observed that performance peaks at eight bins, with both methods notably weaker at four bins. Increasing the bin count beyond eight results in only marginal accuracy changes. At 32 bins, BHA–MaxST briefly outperforms PSSI–MaxST, whereas at 64 bins PSSI–MaxST regains a slight edge. These observations indicate that finer histogram quantization yields diminishing returns.

\subsection{Challenging cases}

As with other approaches that rely on low‑level segmentation, the nodes of our graph $G$ are pixel‑segments generated by those methods. Consequently, our framework inherits the same challenges faced by low-level segmentation methods. Though, we found that SLIC and MeanShift are quite robust to similar foreground and background settings and using well-placed scribbles can give optimal results. Although PSSI–MaxST reduces the number of user scribbles required, accurate and well‑placed scribbles remain essential for achieving high‑quality segmentation.

\section{Conclusion}
\label{Conclusion}

To address the limitations of existing graph-based interactive image segmentation techniques such as OneCut, SSNCut, and AMOE, which arise primarily due to suboptimal similarity measures, we introduced a novel similarity metric called the Pixel Segment Similarity Index (PSSI). This metric is used for assigning edge weights between pixel-segments obtained via MeanShift segmentation. The segmentation is carried out by partitioning the constructed graph using a Maximum Spanning Tree (MaxST), which effectively captures strongly connected local structures. The proposed PSSI metric incorporates color histogram similarities while taking into account local neighborhood consistency, resulting in more accurate segmentation boundaries. Experimental evaluations demonstrate that the proposed PSSI–MaxST framework outperforms current graph-based interactive image segmentation methods in terms of segmentation quality, as measured by Jaccard Index, $F_1$ score, and Mean Error (ME).

A key strength of the method lies in its interactive nature, where user input via scribbles directly influences the segmentation outcome. This makes it particularly suitable for high-precision applications where user control is essential. 
While the proposed method is currently designed for binary segmentation, it provides a strong foundation for extension to graph-based multi-object segmentation. Future work will explore generalizations using advanced graph-theoretic techniques, such as hierarchical MST cuts, to handle multiple object partitions within an image.

\bibliography{refs}

\section*{Biographies}
\begin{IEEEbiography}[{\includegraphics[width=1in,height=1in,clip,keepaspectratio]{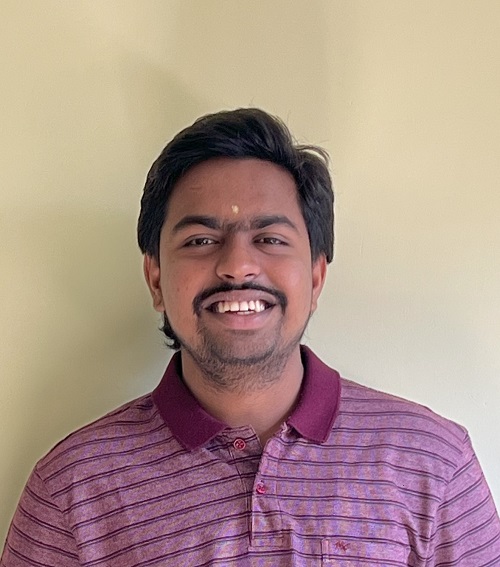}}]{Kaustubh Shivshankar Shejole}
received the bachelor's degree in Computer Science and Engineering from VNIT Nagpur in 2024. He is currently a Ph.D. scholar at IIT Bombay, India. His research interests include Image Processing, Graph Based Learning, Natural Language Processing, Machine Learning, etc.
\end{IEEEbiography}

\begin{IEEEbiography}[{\includegraphics[width=1in,height=1in,clip,keepaspectratio]{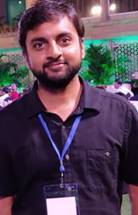}}]{Gaurav Mishra}
received the PhD degree in Computer Science and Engineering from the Indian Institute of Information Technology, Design and Manufacturing, Jabalpur, MP, India. He is currently working as an assistant professor at VNIT, Nagpur. His research interests include Machine Learning, Graph-based Data Mining, and Image Analysis.
\end{IEEEbiography}

\end{document}